\documentclass{article}

\usepackage{arxiv}

\usepackage[utf8]{inputenc} 
\usepackage[T1]{fontenc}    
\usepackage{hyperref}       
\usepackage{url}            
\usepackage{booktabs}       
\usepackage{amsfonts}       
\usepackage{nicefrac}       
\usepackage{microtype}      
\usepackage{lipsum}         
\usepackage{graphicx}
\usepackage{natbib}
\usepackage{doi}
\usepackage{tikz}  
\usetikzlibrary{arrows, positioning}  
\usepackage{subcaption}
\usepackage{multirow}
\usepackage{amsmath}
\usepackage{cleveref}       

\title{\LARGE 
\textbf{FairCauseSyn}: Towards Causally Fair LLM-augmented Synthetic Data Generation 
}

\author{
  Nitish Nagesh \\
  University of California, Irvine \\
  \texttt{nnagesh1@uci.edu}
  \And
  Ziyu Wang \\
  University of California, Irvine \\
  \texttt{ziyuw31@uci.edu}
  \And
  Amir M. Rahmani \\
  University of California, Irvine \\
  \texttt{amirr1@uci.edu}
}

\renewcommand{\shorttitle}{\textit{arXiv} Template}

\hypersetup{
    pdftitle={FairCauseSyn: Towards Causally Fair LLM-Augmented Synthetic Data Generation},
    pdfauthor={Nitish Nagesh, Ziyu Wang, Amir M. Rahmani},
    pdfkeywords={Synthetic Data, Causal Fairness, LLMs, Health Equity}
}

\begin{document}
\maketitle

\begin{abstract}
Synthetic data generation creates data based on real-world data using generative models. In health applications, generating high-quality data while maintaining fairness for sensitive attributes is essential for equitable outcomes. Existing GAN-based and LLM-based methods focus on counterfactual fairness and are primarily applied in finance and legal domains. Causal fairness provides a more comprehensive evaluation framework by preserving causal structure, but current synthetic data generation methods do not address it in health settings. To fill this gap, we develop the first LLM-augmented synthetic data generation method to enhance causal fairness using real-world tabular health data. Our generated data deviates by less than 10\% from real data on causal fairness metrics. When trained on causally fair predictors, synthetic data reduces bias on the sensitive attribute by 70\% compared to real data. This work improves access to fair synthetic data, supporting equitable health research and healthcare delivery.
\end{abstract}


\section{Introduction}
Synthetic data generation is increasingly being adopted across various domains to facilitate downstream prediction tasks \cite{jordon2022synthetic}. 
Since synthetic data is often grounded in real-world datasets, biases inherent in the original data can persist, raising fairness concerns that may negatively impact critical applications such as healthcare \cite{mehrabi2021survey}. 
To ensure equitable health outcomes, it is imperative that synthetic data is generated in a manner that accounts for and mitigates bias \cite{bhanot2021problem}.

High-quality synthetic data generation depends on preserving the statistical properties of the original data. 
Deep generative models such as CTGAN and TVAE \cite{xu2019modeling} have enabled tabular synthetic data generation, and more recently, diffusion-based methods like TabDDPM \cite{kotelnikov2023tabddpm} have further improved data fidelity. 
Moreover, the advent of large language models (LLMs) has introduced new paradigms in synthetic tabular data generation, as demonstrated by REaLTabFormer \cite{solatorio2023realtabformer} and BeGReaT \cite{borisov2022language}.

Despite advancements in synthetic data generation, fairness remains a critical challenge. 
Existing fairness-aware synthetic data generation methods primarily focus on counterfactual fairness, which ensures that protected attributes do not influence predictions under predefined fairness constraints.
For instance, TabFairGAN \cite{rajabi2022tabfairgan} extends CTGAN and TVAE by incorporating fairness constraints to mitigate statistical and demographic parity violations.
Counterfactual fairness, introduced by \cite{kusner2017counterfactual}, provides a more nuanced fairness notion by ensuring that outcomes remain unchanged under hypothetical interventions on protected attributes.
This concept was extended through path-specific counterfactual fairness \cite{chiappa2019path}, forming the basis for fairness-aware generative models such as FairGAN \cite{xu2018fairgan} and CFGAN \cite{xu2019cfan}. 
Although these models address fairness concerns in synthetic data generation, they primarily focus on counterfactual fairness and often fail to satisfy multiple fairness constraints, particularly in classification tasks.

In contrast, causal fairness provides a broader framework by explicitly modeling the underlying causal relationships in data.
It has been formalized through causal fairness metrics such as total effect (TV), direct effect (DE), indirect effect (IE), and spurious effect (SE) \cite{zhang2018fairness, nabi2019learning, plevcko2024causal}.
A key distinction between counterfactual fairness and causal fairness is highlighted by \cite{schroder2023causal}, which underscores that counterfactual fairness methods do not necessarily account for causal pathways leading to biased outcomes.
Causal fairness metrics aim to disentangle and quantify sources of bias beyond observational constraints, offering a more robust fairness framework.

Existing fairness-aware synthetic data generation methods have primarily been evaluated on finance and legal domains \cite{van2021decaf} with limited applications to health datasets.
Existing LLM-based synthetic data generation methods \cite{subahmitigating} are designed keeping to satisfy counterfactual fairness constraints.  To the best of our knowledge, we are the first to develop an LLM-augmented synthetic data generation pipeline that explicitly enhances causal fairness for health datasets.

Our contributions are as follows:
\begin{enumerate}
    \item We propose an LLM-augmented synthetic data generation framework that enforces causal fairness constraints.
    \item We evaluate both real-world health data and generated synthetic data using causal fairness metrics.
    \item We train prediction models with and without fairness constraints to analyze the decomposition of causal fairness metrics.
    \item We demonstrate that our LLM-generated synthetic data, trained with causally fair predictors, achieves near-zero direct and indirect effects, along with lower spurious effects compared to real-world health data.
\end{enumerate}

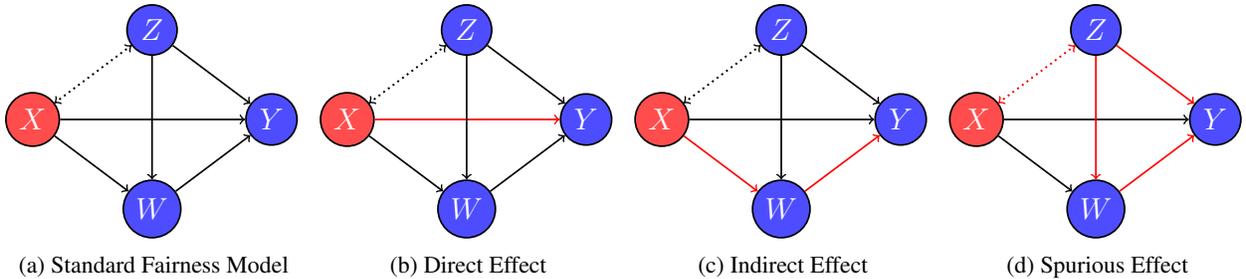
\begin{figure*}[ht]
\centering
\begin{minipage}[t]{0.24\textwidth}
\centering
\resizebox{\columnwidth}{!}{%
\begin{tikzpicture}[node distance=0.5cm and 0.5cm, every node/.style={draw, circle, thick, minimum size=0.8cm}]
\node[fill=red!70, text=white] (X) at (0, 0) {\Large $X$};
\node[fill=blue!70, text=white] (Z) at (2, 1.5) {\Large $Z$};
\node[fill=blue!70, text=white] (W) at (2, -1.5) {\Large $W$};
\node[fill=blue!70, text=white] (Y) at (4, 0) {\Large $Y$};
\draw[->, dotted, thick] (X) -- (Z);
\draw[->, thick] (X) -- (W);
\draw[->, thick] (X) -- (Y);
\draw[->, thick] (Z) -- (Y);
\draw[->, thick] (W) -- (Y);
\draw[->, thick] (Z) -- (W);
\draw[->, dotted, thick] (Z) -- (X);
\end{tikzpicture}
}
\subcaption{Standard Fairness Model}
\label{fig:sfm}
\end{minipage}
\hfill
\begin{minipage}[t]{0.24\textwidth}
\centering
\resizebox{\columnwidth}{!}{%
\begin{tikzpicture}[node distance=0.5cm and 0.5cm, every node/.style={draw, circle, thick, minimum size=0.8cm}]
\node[fill=red!70, text=white] (X) at (0, 0) {\Large $X$};
\node[fill=blue!70, text=white] (Z) at (2, 1.5) {\Large $Z$};
\node[fill=blue!70, text=white] (W) at (2, -1.5) {\Large $W$};
\node[fill=blue!70, text=white] (Y) at (4, 0) {\Large $Y$};
\draw[->, dotted, thick] (X) -- (Z);
\draw[->, thick] (X) -- (W);
\draw[->, thick, draw=red] (X) -- (Y);
\draw[->, thick] (Z) -- (Y);
\draw[->, thick] (W) -- (Y);
\draw[->, thick] (Z) -- (W);
\draw[->, dotted, thick] (Z) -- (X);
\end{tikzpicture}
}
\subcaption{Direct Effect}
\label{fig:direct}
\end{minipage}
\hfill
\begin{minipage}[t]{0.24\textwidth}
\centering
\resizebox{\columnwidth}{!}{%
\begin{tikzpicture}[node distance=0.5cm and 0.5cm, every node/.style={draw, circle, thick, minimum size=0.8cm}]
\node[fill=red!70, text=white] (X) at (0, 0) {\Large $X$};
\node[fill=blue!70, text=white] (Z) at (2, 1.5) {\Large $Z$};
\node[fill=blue!70, text=white] (W) at (2, -1.5) {\Large $W$};
\node[fill=blue!70, text=white] (Y) at (4, 0) {\Large $Y$};
\draw[->, dotted, thick] (X) -- (Z);
\draw[->, thick] (X) -- (Y);
\draw[->, thick, draw=red] (X) -- (W);
\draw[->, thick, draw=red] (W) -- (Y);
\draw[->, thick] (Z) -- (Y);
\draw[->, thick] (Z) -- (W);
\draw[->, dotted, thick] (Z) -- (X);
\end{tikzpicture}
}
\subcaption{Indirect Effect}
\label{fig:indirect}
\end{minipage}
\hfill
\begin{minipage}[t]{0.24\textwidth}
\centering
\resizebox{\columnwidth}{!}{%
\begin{tikzpicture}[node distance=0.5cm and 0.5cm, every node/.style={draw, circle, thick, minimum size=0.8cm}]
\node[fill=red!70, text=white] (X) at (0, 0) {\Large $X$};
\node[fill=blue!70, text=white] (Z) at (2, 1.5) {\Large $Z$};
\node[fill=blue!70, text=white] (W) at (2, -1.5) {\Large $W$};
\node[fill=blue!70, text=white] (Y) at (4, 0) {\Large $Y$};
\draw[->, dotted, thick, draw=red] (X) -- (Z);
\draw[->, thick] (X) -- (Y);
\draw[->, thick] (X) -- (W);
\draw[->, thick, draw=red] (W) -- (Y);
\draw[->, thick, draw=red] (Z) -- (Y);
\draw[->, thick, draw=red] (Z) -- (W);
\draw[->, dotted, thick, draw=red] (Z) -- (X);
\end{tikzpicture}
}
\subcaption{Spurious Effect}
\label{fig:spurious}
\end{minipage}

\caption{Standard Fairness Model (SFM) and different causal effects.
'X' refers to the protected attribute, 'Z' is the demographic variable, 'W' are mediators, 'Y' is the outcome variable
(a) Standard Fairness Model, (b) Direct Effect - effect of protected attribute on outcome, (c) Indirect Effect - effect of protected attributes and mediators on outcome, and (d) Spurious Effect - effect of confounders and mediators.}
\label{fig:sfm_all}
\end{figure*}

\section{Methodology}

To address the challenges of fairness in decision-making processes, we introduce the \textit{FairCauseSyn} pipeline, a comprehensive framework for understanding, quantifying, and mitigating discrimination in data-driven systems. This pipeline integrates concepts from causal inference to model and evaluate fairness rigorously, leveraging structural causal models (SCMs) and the standard fairness model (SFM) to decompose and measure causal effects. In addition, the framework employs advanced synthetic data generation techniques, supported by large language models (LLMs), to produce fair and representative data for downstream applications. The following subsections detail the theoretical foundations, fairness evaluation metrics, and the synthetic data generation process that form the backbone of this framework.

\begin{figure*}[ht]
    \centering
    \includegraphics[width=\textwidth]{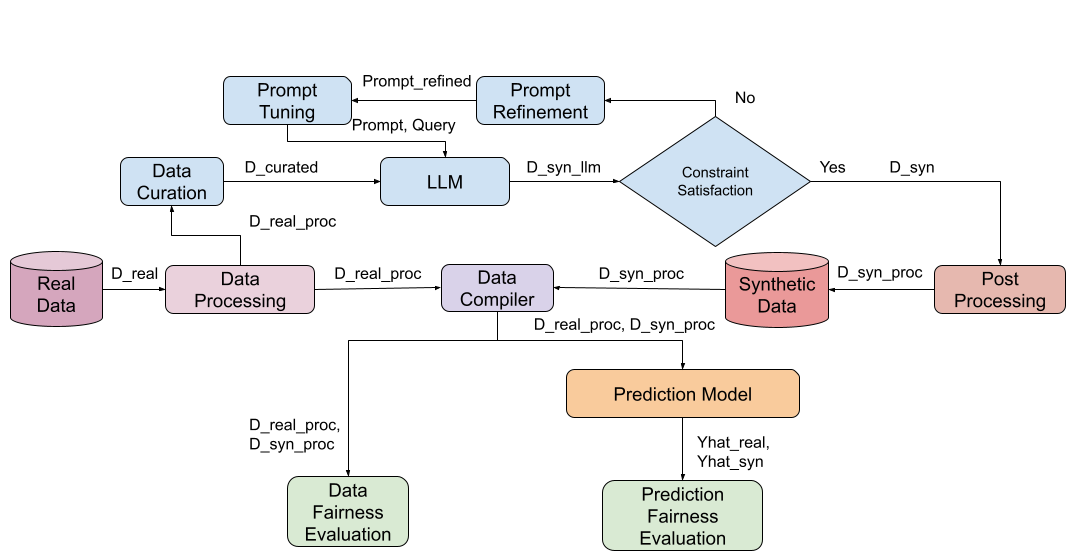} 
    \caption{FairCauseSyn generation and evaluation framework}
    \label{fig:fair_sdg}
\end{figure*}

\subsection{Structural Causal Model}

A structural causal model (SCM) \cite{plevcko2024causal} is a 4-tuple $\langle V, U, \mathcal{F}, P(u) \rangle$, where $U$ represents a unobserved variables,  $V = \{V_1, \dots, V_n\}$ is a collection of observed variables, $\mathcal{F} = \{f_1, \dots, f_n\}$ is used to estimate $V$, $v_i \leftarrow f_i(\text{pa}(v_i), u_i)$, where $\text{pa}(V_i) \subseteq V \setminus V_i$ and $U_i \subseteq U$ are the functional arguments of $f_i$ and $P(u)$ is a distribution over the unobserved variables $U$.

The structural equations $\mathcal{F}$ encode the deterministic causal relationships between variables, while $P(U)$ introduces stochasticity into the model through the unobserved variables. Together, these components enable reasoning about causal relationships, interventions, and counterfactuals within the system.

\subsection{Standard Fairness Model}
The \textit{standard fairness model} (SFM) \cite{plevcko2024causal} is a way of representing relationship between observed variables and evaluate causal fairness. As shown in Fig.~\ref{fig:sfm}, $X$ represents the protected or sensitive attribute (sex, race), $Z$ captures observed confounders (demographic attributes such as age), $W$ indicates mediators (influenced by sensitive attribute and afffecting outcome) and $Y$ is the outcome of interest (e.g., clinical decisions).

SFM decomposes the causal relationship between the protected attribute ($X$) and the outcome ($Y$) into three key effects:
\begin{enumerate}
    \item \textbf{Direct Effect (DE):} The direct causal influence of $X$ on $Y$ ($X \rightarrow Y$),
    \item \textbf{Indirect Effect (IE):} The mediated causal influence of $X$ on $Y$ through $W$ ($X \rightarrow W \rightarrow Y$),
    \item \textbf{Spurious Effect (SE):} Confounding influences arising from $Z$, including paths such as $Z \rightarrow W \rightarrow Y$, $Z \rightarrow Y$, and bidirectional paths like $X \leftrightarrow Z$.
\end{enumerate}
The different effects are captured in Figure~\ref{fig:sfm_all}

\subsection{Causal Fairness Metrics}

To assess fairness, the SFM leverages path-specific causal fairness measures \cite{schroder2023causal}, focusing on the discrimination caused by specific causal pathways.
Discrimination is formalized as comparison of $X = x_0$ compared to $X = x_1$ for pathway conditioned on $X = x$, i.e.,
\begin{align}
    \text{DE}_{x_0, x_1}(y \mid x) &:= P(y_{x_1, m_{x_0}} \mid x) - P(y_{x_0} \mid x), \tag{1} \\
    \text{IE}_{x_0, x_1}(y \mid x) &:= P(y_{x_0, m_{x_1}} \mid x) - P(y_{x_0} \mid x), \tag{2} \\
    \text{SE}_{x_0, x_1}(y) &:= P(y_{x_0} \mid x_1) - P(y \mid x_0). \tag{3}
\end{align}

The total variation, i.e., the overall discrimination of individuals with $X = x_0$ compared to $X = x_1$, can then be explained as
\[
TV_{x_0, x_1} = \text{DE}_{x_0, x_1}(y \mid x_0) - \text{IE}_{x_1, x_0}(y \mid x_0) - \text{SE}_{x_0, x_1}(y)
\]

Causal fairness is achieved when $\text{DE}_{x_0, x_1}(y \mid x)$, $\text{IE}_{x_0, x_1}(y \mid x)$, $\text{SE}_{x_0, x_1}(y)$ are zero or close to zero.

\subsection{Synthetic Data Generation Framework}

The \textit{FairCauseSyn} framework is designed to generate synthetic data that satisfies fairness constraints while preserving the utility of the original dataset. As illustrated in Fig.~\ref{fig:fair_sdg}, the pipeline begins with real clinical data \( D_{\text{real}} \), which is preprocessed into a structured format \( D_{\text{real\_proc}} \) suitable for modeling.

The processed data serves two purposes: (1) it provides a basis for fairness evaluation using causal decomposition metrics such as Total Variation (TV), Direct Effect (DE), Indirect Effect (IE), and Spurious Effect (SE) \cite{plevcko2024causal}; and (2) it guides the generation of synthetic data through large language models (LLMs).

Synthetic data generation is performed via a prompt-driven process. Curated subsets of \( D_{\text{real\_proc}} \) inform prompt construction through techniques such as self-consistency \cite{wang2022self}, in-context learning \cite{dong2024survey}, and schema-guided generation \cite{zhang2023sgp}. These prompts are used to query the LLM, producing candidate synthetic samples \( D_{\text{syn\_llm}} \). A constraint satisfaction module evaluates each batch for adherence to fairness and data fidelity requirements. If constraints are unmet, iterative prompt refinement and data adaptation steps are invoked. This loop continues until a final synthetic dataset \( D_{\text{syn}} \) is obtained, followed by post-processing for downstream compatibility.

To assess the utility and fairness of the synthetic data, two predictive models are employed: (1) a baseline model without fairness constraints and (2) a causally fair prediction model based on the FairAdapt framework \cite{plevcko2024fairadapt}. Both models are trained and evaluated on real and synthetic data, enabling comparative analysis of fairness preservation and predictive performance.

\begin{table*}[ht]
\centering
\renewcommand{\arraystretch}{1.5}  
\resizebox{0.8\textwidth}{!}{%
\begin{tabular}{|l|l|c|c|c|c|}
\hline
\textbf{Fairness Evaluation} & \textbf{Dataset} & \textbf{Total Variation (TV) $\downarrow$}   & \textbf{Direct Effect (DE) $\downarrow$} & \textbf{Indirect Effect (IE) $\downarrow$} & \textbf{Spurious Effect (SE) $\downarrow$} \\ \hline
\multirow{2}{*}{Data Fairness} 
    & Real      & $-0.0121 \pm 0.0537$ & $-0.0477 \pm 0.0026$ & $-0.0472 \pm 0.0068$ & $0.0116 \pm 0.0556$ \\ \cline{2-6} 
    & Synthetic & $-0.0492 \pm 0.0571$ & $-0.0429 \pm 0.0043$ & $-0.0002 \pm 0.0072$ & $0.0064 \pm 0.0580$ \\ \hline
\multirow{2}{*}{Prediction Model w/o Fairness Constraints} 
    & Real      & $0.0130 \pm 0.0458$ & $0.0060 \pm 0.0022$ & $-0.0163 \pm 0.0074$ & $0.0093 \pm 0.0477$ \\ \cline{2-6} 
    & Synthetic & $-0.0433 \pm 0.0550$ & $-0.0287 \pm 0.0037$ & $-0.0217 \pm 0.0117$ & $0.0363 \pm 0.0552$ \\ \hline
\multirow{2}{*}{Causally Fair Prediction Model} 
    & Real      & $0.0248 \pm 0.0631$ & $-0.0070 \pm 0.0016$ & $-0.0538 \pm 0.0054$ & $0.0219 \pm 0.0637$ \\ \cline{2-6} 
    & Synthetic & $0.0003 \pm 0.0568$ & \textbf{$-0.0020 \pm 0.0030$} & \textbf{$0.0076 \pm 0.0108$} & $-0.0099 \pm 0.0556$ \\ \hline
\end{tabular}%
}
\caption{Causal fairness metrics (Total Variation, Direct Effect, Indirect Effect, Spurious Effect) for real and synthetic datasets under three modeling conditions: raw data, prediction without fairness constraints, and prediction with causally fair modeling. Results are reported as mean $\pm$ standard deviation.}
\label{tab:fairness_results}
\end{table*}

\section{Experiments and Results}

The purpose of this section is to evaluate the effectiveness of the \textit{FairCauseSyn} framework in addressing fairness concerns in healthcare decision-making systems. We conduct a case study on heart failure clinical records to demonstrate how structural causal modeling and fairness analysis can be applied to real-world clinical datasets. This case study highlights the utility of synthetic data generation and fairness evaluation in ensuring equitable outcomes. 

\subsection{Dataset}

This study utilizes the Heart Failure Clinical Records dataset \cite{chicco2020machine}, comprising 299 patients who underwent follow-up care for heart failure. The dataset includes demographic attributes (age, sex), clinical indicators (anaemia, blood pressure, creatinine phosphokinase levels, ejection fraction, serum sodium and creatinine, platelet count), and comorbidities (diabetes, smoking). Outcome variables include follow-up duration and survival status, denoted by a binary variable.

This dataset was selected due to its clinical relevance and suitability for evaluating fairness-aware modeling. Specifically, disparities in outcomes related to age and gender provide a valuable context for applying structural causal modeling and assessing synthetic data fairness interventions.

\subsection{Standard Fairness Model for Dataset}

We apply the Standard Fairness Model (SFM) to analyze fairness in the heart failure clinical records dataset \cite{chicco2020machine}, focusing on understanding causal relationships and fairness in survival outcomes. The SFM provides a structured and interpretable approach to identify and decompose the direct, mediated, and confounded influences of sensitive attributes, enabling targeted interventions to ensure fairness in healthcare decision-making systems.

In this analysis, the protected attribute (\(X\)) is the patient’s sex, while the demographic variable (\(Z\)) represents the patient’s age. Clinical features such as anaemia status, diabetes status, ejection fraction, blood pressure status, serum sodium level, serum creatinine level, smoking status, follow-up period, and platelet count are considered mediators (\(W\)). The outcome (\(Y\)) is the survival status indicating mortality.

Using the SFM, we analyze the following causal pathways to quantify fairness:
\begin{enumerate}
    \item \textbf{Direct Effect (DE):} This pathway (\(X \rightarrow Y\)) measures the direct influence of gender on survival outcomes, reflecting any direct discrimination.
    \item \textbf{Indirect Effect (IE):} The mediated path (\(X \rightarrow W \rightarrow Y\)) captures gendered differences in clinical mediators that influence survival outcomes, reflecting indirect discrimination.
    \item \textbf{Spurious Effect (SE):} Confounding pathways (\(Z \rightarrow W \rightarrow Y\), \(Z \rightarrow Y\), and \(X \leftrightarrow Z\)) capture correlations between age, clinical mediators, and outcomes that may spuriously inflate or distort causal effects.
\end{enumerate}

The SFM allows us to systematically evaluate these pathways, enabling a detailed understanding of how unfairness arises in clinical predictions and guiding the development of interventions to mitigate it. This analysis demonstrates the utility of causal models in uncovering biases in complex, real-world healthcare datasets.

\subsection{Framework Modules}

We now describe the implementation of each component in the FairCauseSyn framework as used in our experiments.

The real dataset \( D_{\text{real}} \), sourced from heart failure clinical records, is preprocessed to yield \( D_{\text{real\_proc}} \). This involves normalization, handling of missing values, and encoding of categorical features. Causal fairness metrics - TV, DE, IE, and SE - are computed following the methodology in \cite{plevcko2024causal}.

In the \textit{Data Curation} phase, a subset of representative and causally relevant examples from \( D_{\text{real\_proc}} \) is extracted to inform prompt construction. Prompt engineering is applied in the \textit{Prompt Tuning} module, followed by iterative \textit{Prompt Refinement} and \textit{Data Adaptation} to ensure fairness constraints are met. The LLM-generated synthetic data \( D_{\text{syn\_llm}} \) is evaluated using constraint checks; if satisfied, the finalized data \( D_{\text{syn}} \) proceeds to the post-processing stage.

Both \( D_{\text{real\_proc}} \) and \( D_{\text{syn}} \) are then used to train two prediction models: a random forest classifier as the baseline model, and a causally fair model using FairAdapt. These models produce predicted outcomes \( \hat{Y}_{\text{real}} \) and \( \hat{Y}_{\text{syn}} \), respectively. Predictions are evaluated using causal fairness decomposition to assess whether the synthetic data supports fair and reliable decision-making in clinical contexts.

\begin{figure*}[ht]
    \centering
    \includegraphics[width=\textwidth]{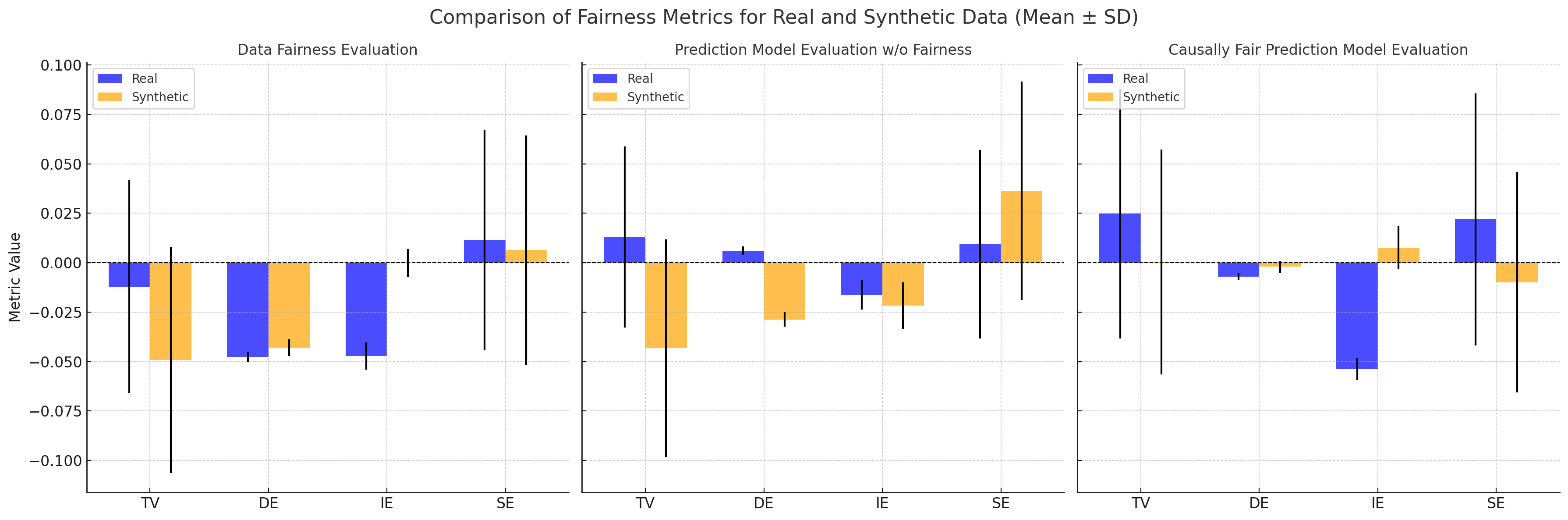} 
    \caption{Comparison of causal fairness metrics - Total Variation (TV), Direct Effect (DE), Indirect Effect (IE), and Spurious Effect (SE) - for real and synthetic data across three evaluation settings: (i) data-level fairness, (ii) prediction without fairness constraints, and (iii) prediction with causally fair modeling. Bars represent mean values with error bars denoting standard deviation (SD).}
    \label{fig:sfm_metrics}
\end{figure*}

\subsection{Results}

The following section presents a comprehensive analysis of the fairness metrics across real and synthetic datasets, evaluating their impact on data fairness, prediction models without fairness constraints, and causally fair prediction models, with a focus on understanding the contributions of Total Variation (TV), Direct Effect (DE), Indirect Effect (IE), and Spurious Effect (SE) to overall fairness. 
The decomposition of fairness metrics across real and synthetic datasets is visualized in Figure~\ref{fig:sfm_metrics}. This figure highlights the relative contributions of DE, IE, and SE to TV under different fairness evaluation models, further corroborating the findings from Table~\ref{tab:fairness_results}.

\subsubsection{Data Fairness Evaluation}

According to \cite{makhlouf2024advancing}, a fairness metric value of zero indicates no discrimination, while positive and negative values reflect bias against or in favor of the protected group, respectively. 

For the real dataset, the Total Variation (TV) was approximately zero ($-0.0121 \pm 0.0537$), indicating overall fairness. However, the Direct Effect (DE) contributed most substantially to the observed disparity ($-0.0477 \pm 0.0026$), suggesting disparate treatment against male patients. The Indirect Effect (IE) and Spurious Effect (SE) were smaller in magnitude ($-0.0472 \pm 0.0068$ and $0.0116 \pm 0.0556$, respectively), with their opposing signs approximately canceling out.

The synthetic dataset exhibited similar trends. The DE remained negative and slightly stronger ($-0.0429 \pm 0.0043$), while the IE was negligible ($-0.0002 \pm 0.0072$). These results indicate that the synthetic data captured key fairness characteristics of the real data. 

Notably, the standard deviations for DE and IE were low, suggesting reliable estimation. In contrast, SE showed substantial variability in both datasets, highlighting ongoing challenges in mitigating spurious associations during synthetic data generation.

\subsubsection{Prediction Model without Fairness Constraints}

The prediction model trained without fairness constraints exhibited observable biases across both the real and synthetic datasets. For the real data, the Total Variation (TV) was slightly positive ($0.0130 \pm 0.0458$), with marginal contributions from the Direct Effect (DE: $0.0060 \pm 0.0022$) and Indirect Effect (IE: $-0.0163 \pm 0.0074$). However, the Spurious Effect (SE: $0.0093 \pm 0.0477$) exhibited substantial variability, indicating instability in the model’s mortality predictions.

In the synthetic data, the TV shifted to a negative value ($-0.0433 \pm 0.0550$), primarily driven by heightened SE variability ($0.0363 \pm 0.0552$) and inverted contributions from DE and IE. The relatively high standard deviations across components in both datasets suggest reduced estimation confidence, likely attributable to data imbalance and the presence of outliers.

\subsubsection{Causally Fair Prediction Model}

The causally fair prediction model is designed to reduce bias by preserving underlying causal structures during training. In the real dataset, the Direct Effect (DE) was significantly reduced ($-0.0070 \pm 0.0016$), indicating a lower level of disparate treatment. However, the Indirect Effect (IE: $-0.0538 \pm 0.0054$) and Spurious Effect (SE: $0.0219 \pm 0.0637$) contributed to a higher Total Variation (TV: $0.0248 \pm 0.0631$). The elevated SE variability suggests that spurious influences were not fully eliminated.

For the synthetic dataset, all three fairness components were further minimized - DE ($-0.0020 \pm 0.0030$), IE ($0.0076 \pm 0.0108$), and SE ($-0.0099 \pm 0.0556$) - demonstrating improved mitigation of disparate treatment and impact. However, the reversal in direction for IE and SE indicates possible differences in causal structure preservation between the datasets.

Although the causally fair model effectively reduces DE and IE contributions, the consistently high standard deviation in SE - especially in synthetic data - emphasizes persistent challenges in controlling spurious effects. These findings suggest that minimizing TV alone is insufficient; robust mitigation of SE remains essential for achieving true causal fairness in predictive modeling.

\section{Conclusion and Future Work}

We introduced an LLM-augmented pipeline for synthetic data generation that embeds causal fairness constraints - addressing a key gap in fairness-aware data synthesis. Our evaluation shows that the generated data closely mirrors real-world health data across fairness metrics, with deviations in Total Variation (TV), Direct Effect (DE), and Indirect Effect (IE) consistently below 10\%. When used to train models under causal fairness constraints, the synthetic data substantially reduced biases in skewed health datasets, with reductions in direct effect exceeding 70\%.

These results demonstrate the potential of LLMs in generating high-quality, causally fair synthetic data and offer a promising direction for equitable health research and policy modeling. Crucially, our approach goes beyond counterfactual fairness by incorporating causal fairness within the data generation process.

Future work will address limitations in controlling spurius effect, which showed higher variability in the synthetic data. We aim to integrate prompt optimization and structured causal modeling to mitigate spurious correlations while maintaining fairness constraints.

\section*{Acknowledgments}
We thank Iman Azimi for his support in brainstorming and positioning our work. \\
This paper was accepted at IEEE EMBC 2025. This is the author's version of the accepted manuscript. The final published version will be available via IEEE Xplore.



\begin{thebibliography}{24}
\providecommand{\natexlab}[1]{#1}
\providecommand{\url}[1]{\texttt{#1}}
\expandafter\ifx\csname urlstyle\endcsname\relax
  \providecommand{\doi}[1]{doi: #1}\else
  \providecommand{\doi}{doi: \begingroup \urlstyle{rm}\Url}\fi

\bibitem[Jordon et~al.(2022)Jordon, Szpruch, Houssiau, Bottarelli, Cherubin, Maple, Cohen, and Weller]{jordon2022synthetic}
James Jordon, Lukasz Szpruch, Florimond Houssiau, Mirko Bottarelli, Giovanni Cherubin, Carsten Maple, Samuel~N Cohen, and Adrian Weller.
\newblock Synthetic data--what, why and how?
\newblock \emph{arXiv preprint arXiv:2205.03257}, 2022.

\bibitem[Mehrabi et~al.(2021)Mehrabi, Morstatter, Saxena, Lerman, and Galstyan]{mehrabi2021survey}
Ninareh Mehrabi, Fred Morstatter, Nripsuta Saxena, Kristina Lerman, and Aram Galstyan.
\newblock A survey on bias and fairness in machine learning.
\newblock \emph{ACM computing surveys (CSUR)}, 54\penalty0 (6):\penalty0 1--35, 2021.

\bibitem[Bhanot et~al.(2021)Bhanot, Qi, Erickson, Guyon, and Bennett]{bhanot2021problem}
Karan Bhanot, Miao Qi, John~S Erickson, Isabelle Guyon, and Kristin~P Bennett.
\newblock The problem of fairness in synthetic healthcare data.
\newblock \emph{Entropy}, 23\penalty0 (9):\penalty0 1165, 2021.

\bibitem[Xu et~al.(2019{\natexlab{a}})Xu, Skoularidou, Cuesta-Infante, and Veeramachaneni]{xu2019modeling}
Lei Xu, Maria Skoularidou, Alfredo Cuesta-Infante, and Kalyan Veeramachaneni.
\newblock Modeling tabular data using conditional gan.
\newblock \emph{Advances in neural information processing systems}, 32, 2019{\natexlab{a}}.

\bibitem[Kotelnikov et~al.(2023)Kotelnikov, Baranchuk, Rubachev, and Babenko]{kotelnikov2023tabddpm}
Akim Kotelnikov, Dmitry Baranchuk, Ivan Rubachev, and Artem Babenko.
\newblock Tabddpm: Modelling tabular data with diffusion models.
\newblock In \emph{International Conference on Machine Learning}, pages 17564--17579. PMLR, 2023.

\bibitem[Solatorio and Dupriez(2023)]{solatorio2023realtabformer}
Aivin~V Solatorio and Olivier Dupriez.
\newblock Realtabformer: Generating realistic relational and tabular data using transformers.
\newblock \emph{arXiv preprint arXiv:2302.02041}, 2023.

\bibitem[Borisov et~al.(2022)Borisov, Se{\ss}ler, Leemann, Pawelczyk, and Kasneci]{borisov2022language}
Vadim Borisov, Kathrin Se{\ss}ler, Tobias Leemann, Martin Pawelczyk, and Gjergji Kasneci.
\newblock Language models are realistic tabular data generators.
\newblock \emph{arXiv preprint arXiv:2210.06280}, 2022.

\bibitem[Rajabi and Garibay(2022)]{rajabi2022tabfairgan}
Amirarsalan Rajabi and Ozlem~Ozmen Garibay.
\newblock Tabfairgan: Fair tabular data generation with generative adversarial networks.
\newblock \emph{Machine Learning and Knowledge Extraction}, 4\penalty0 (2):\penalty0 488--501, 2022.

\bibitem[Kusner et~al.(2017)Kusner, Loftus, Russell, and Silva]{kusner2017counterfactual}
Matt~J Kusner, Joshua Loftus, Chris Russell, and Ricardo Silva.
\newblock Counterfactual fairness.
\newblock \emph{Advances in neural information processing systems}, 30, 2017.

\bibitem[Chiappa(2019)]{chiappa2019path}
Silvia Chiappa.
\newblock Path-specific counterfactual fairness.
\newblock In \emph{Proceedings of the AAAI conference on artificial intelligence}, volume~33, pages 7801--7808, 2019.

\bibitem[Xu et~al.(2018)Xu, Yuan, Zhang, and Wu]{xu2018fairgan}
Depeng Xu, Shuhan Yuan, Lu~Zhang, and Xintao Wu.
\newblock Fairgan: Fairness-aware generative adversarial networks.
\newblock In \emph{2018 IEEE international conference on big data (big data)}, pages 570--575. IEEE, 2018.

\bibitem[Xu et~al.(2019{\natexlab{b}})Xu, Wu, Yuan, Zhang, and Wu]{xu2019cfan}
Depeng Xu, Yongkai Wu, Shuhan Yuan, Lu~Zhang, and Xintao Wu.
\newblock Achieving causal fairness through generative adversarial networks.
\newblock In \emph{Proceedings of the Twenty-Eighth International Joint Conference on Artificial Intelligence}, 2019{\natexlab{b}}.

\bibitem[Zhang and Bareinboim(2018)]{zhang2018fairness}
Junzhe Zhang and Elias Bareinboim.
\newblock Fairness in decision-making—the causal explanation formula.
\newblock In \emph{Proceedings of the AAAI Conference on Artificial Intelligence}, volume~32, 2018.

\bibitem[Nabi et~al.(2019)Nabi, Malinsky, and Shpitser]{nabi2019learning}
Razieh Nabi, Daniel Malinsky, and Ilya Shpitser.
\newblock Learning optimal fair policies.
\newblock In \emph{International Conference on Machine Learning}, pages 4674--4682. PMLR, 2019.

\bibitem[Ple{\v{c}}ko et~al.(2024{\natexlab{a}})Ple{\v{c}}ko, Bareinboim, et~al.]{plevcko2024causal}
Drago Ple{\v{c}}ko, Elias Bareinboim, et~al.
\newblock Causal fairness analysis: a causal toolkit for fair machine learning.
\newblock \emph{Foundations and Trends{\textregistered} in Machine Learning}, 17\penalty0 (3):\penalty0 304--589, 2024{\natexlab{a}}.

\bibitem[Schr{\"o}der et~al.(2023)Schr{\"o}der, Frauen, and Feuerriegel]{schroder2023causal}
Maresa Schr{\"o}der, Dennis Frauen, and Stefan Feuerriegel.
\newblock Causal fairness under unobserved confounding: a neural sensitivity framework.
\newblock \emph{arXiv preprint arXiv:2311.18460}, 2023.

\bibitem[Van~Breugel et~al.(2021)Van~Breugel, Kyono, Berrevoets, and Van~der Schaar]{van2021decaf}
Boris Van~Breugel, Trent Kyono, Jeroen Berrevoets, and Mihaela Van~der Schaar.
\newblock Decaf: Generating fair synthetic data using causally-aware generative networks.
\newblock \emph{Advances in Neural Information Processing Systems}, 34:\penalty0 22221--22233, 2021.

\bibitem[Subah()]{subahmitigating}
Faria~Zarin Subah.
\newblock Mitigating and assessing bias and fairness in large language model-generated synthetic tabular data.

\bibitem[Wang et~al.(2022)Wang, Wei, Schuurmans, Le, Chi, Narang, Chowdhery, and Zhou]{wang2022self}
Xuezhi Wang, Jason Wei, Dale Schuurmans, Quoc Le, Ed~Chi, Sharan Narang, Aakanksha Chowdhery, and Denny Zhou.
\newblock Self-consistency improves chain of thought reasoning in language models.
\newblock \emph{arXiv preprint arXiv:2203.11171}, 2022.

\bibitem[Dong et~al.(2024)Dong, Li, Dai, Zheng, Ma, Li, Xia, Xu, Wu, Chang, et~al.]{dong2024survey}
Qingxiu Dong, Lei Li, Damai Dai, Ce~Zheng, Jingyuan Ma, Rui Li, Heming Xia, Jingjing Xu, Zhiyong Wu, Baobao Chang, et~al.
\newblock A survey on in-context learning.
\newblock In \emph{Proceedings of the 2024 Conference on Empirical Methods in Natural Language Processing}, pages 1107--1128, 2024.

\bibitem[Zhang et~al.(2023)Zhang, Peng, Li, Zhou, and Meng]{zhang2023sgp}
Xiaoying Zhang, Baolin Peng, Kun Li, Jingyan Zhou, and Helen Meng.
\newblock Sgp-tod: Building task bots effortlessly via schema-guided llm prompting.
\newblock \emph{arXiv preprint arXiv:2305.09067}, 2023.

\bibitem[Ple{\v{c}}ko et~al.(2024{\natexlab{b}})Ple{\v{c}}ko, Bennett, and Meinshausen]{plevcko2024fairadapt}
Drago Ple{\v{c}}ko, Nicolas Bennett, and Nicolai Meinshausen.
\newblock fairadapt: Causal reasoning for fair data preprocessing.
\newblock \emph{Journal of Statistical Software}, 110:\penalty0 1--35, 2024{\natexlab{b}}.

\bibitem[Chicco and Jurman(2020)]{chicco2020machine}
Davide Chicco and Giuseppe Jurman.
\newblock Machine learning can predict survival of patients with heart failure from serum creatinine and ejection fraction alone.
\newblock \emph{BMC medical informatics and decision making}, 20:\penalty0 1--16, 2020.

\bibitem[Makhlouf(2024)]{makhlouf2024advancing}
Karima Makhlouf.
\newblock \emph{Advancing Ethical and Responsible AI: Exploring Fairness, Privacy, and Explainability through Causal Perspectives}.
\newblock PhD thesis, {\'E}cole polytechnique, 2024.

\end{thebibliography}





\end{document}